\documentclass[11pt]{article}

\usepackage[preprint]{acl}

\usepackage{times}
\usepackage{latexsym}

\usepackage[T1]{fontenc}

\usepackage[utf8]{inputenc}

\usepackage{microtype}

\usepackage{inconsolata}

\usepackage{graphicx}
\usepackage{booktabs}
\usepackage{multirow}
\usepackage{array}
\usepackage{tabularx}
\usepackage{xcolor}
\usepackage{xspace}
\usepackage{amsmath} 
\usepackage{amssymb}

\usepackage[ruled,vlined,linesnumbered]{algorithm2e}
\usepackage{float}
\usepackage{caption}
\usepackage{subcaption}
\usepackage{hyperref, enumitem}
\usepackage[dvipsnames]{xcolor}
\usepackage[most]{tcolorbox}
\usepackage{tabularx}
\usepackage[table]{xcolor}
\usepackage{soul}
\setitemize[1]{itemsep=0pt,partopsep=0pt,parsep=0pt,topsep=0pt,leftmargin=12pt}
\newcommand{\sysName}{\textsc{Med-CoReasoner}\xspace}
\newcommand{\dataName}{\textit{MMed-Reason}\xspace}
\newcommand{\evalName}{MultiMed-X\xspace}

\definecolor{theme_green}{HTML}{CDE2E8}
\definecolor{theme_red}{HTML}{F59790}
\definecolor{theme_blue}{HTML}{C8D4E9}
\definecolor{theme_orange}{HTML}{F5DBB6}
\definecolor{theme_purple}{HTML}{DACFE5}
\definecolor{theme_darkpurple}{HTML}{9EAAD1}

\title{\sysName: Reducing Language Disparities in Medical Reasoning via Language-Informed Co-Reasoning}

\author{First Author \\
  Affiliation / Address line 1 \\
  Affiliation / Address line 2 \\
  Affiliation / Address line 3 \\
  \texttt{email@domain} \\\And
  Second Author \\
  Affiliation / Address line 1 \\
  Affiliation / Address line 2 \\
  Affiliation / Address line 3 \\
  \texttt{email@domain} \\}

\author{
 \textbf{Fan Gao\textsuperscript{1}},
 \textbf{Sherry T. Tong\textsuperscript{1}},
 \textbf{Jiwoong Sohn\textsuperscript{2}},
 \textbf{Jiahao Huang\textsuperscript{1,3}},\\
 \textbf{Junfeng Jiang\textsuperscript{3}},
 \textbf{Ding Xia\textsuperscript{1}},
 \textbf{Piyalitt Ittichaiwong\textsuperscript{4}},\\
 \textbf{Kanyakorn Veerakanjana\textsuperscript{4}},
 \textbf{Hyunjae Kim\textsuperscript{5}},
 \textbf{Qingyu Chen\textsuperscript{5}},\\
 \textbf{Edison Marrese Taylor\textsuperscript{1}},
 \textbf{Kazuma Kobayashi\textsuperscript{3}},
 \textbf{Akiko Aizawa\textsuperscript{3}},
 \textbf{Irene Li\textsuperscript{1\ensuremath{\dagger}}}\\
 \textsuperscript{1}The University of Tokyo,
 \textsuperscript{2}ETH Zürich,\\
 \textsuperscript{3}National Institute of Informatics,
 \textsuperscript{4}Siriraj Informatics and Data Innovation Center,\\
 \textsuperscript{5}Yale University
\\
 \textsuperscript{\ensuremath{\dagger}}\small{
   \textbf{Correspondence Author:} \href{irene.li@weblab.t.u-tokyo.ac.jp}{irene.li@weblab.t.u-tokyo.ac.jp}
 }
}

\begin{document}
\maketitle

\begin{abstract}
While reasoning-enhanced large language models perform strongly on English medical tasks, a persistent multilingual gap remains, with substantially weaker reasoning in local languages, limiting equitable global medical deployment. To bridge this gap, we introduce \sysName\footnote{\url{https://github.com/astridesa/Med-CoReasoner}}, a language-informed co-reasoning framework that elicits parallel English and local-language reasoning, abstracts them into structured concepts, and integrates local clinical knowledge into an English logical scaffold via concept-level alignment and retrieval. This design combines the structural robustness of English reasoning with the practice-grounded expertise encoded in local languages. To evaluate multilingual medical reasoning beyond multiple-choice settings, we construct \evalName\footnote{\url{https://huggingface.co/datasets/li-lab/MultiMed-X}}, a benchmark covering seven languages with expert-annotated long-form question answering and natural language inference tasks, comprising 350 instances per language. Experiments across three benchmarks show that \sysName improves multilingual reasoning performance by an average of 5\%, with particularly substantial gains in low-resource languages. Moreover, model distillation and expert evaluation analysis further confirm that \sysName produces clinically sound and culturally grounded reasoning traces.
\end{abstract}


\section{Introduction}
\begin{figure}
    \centering
    \includegraphics[width=\linewidth]{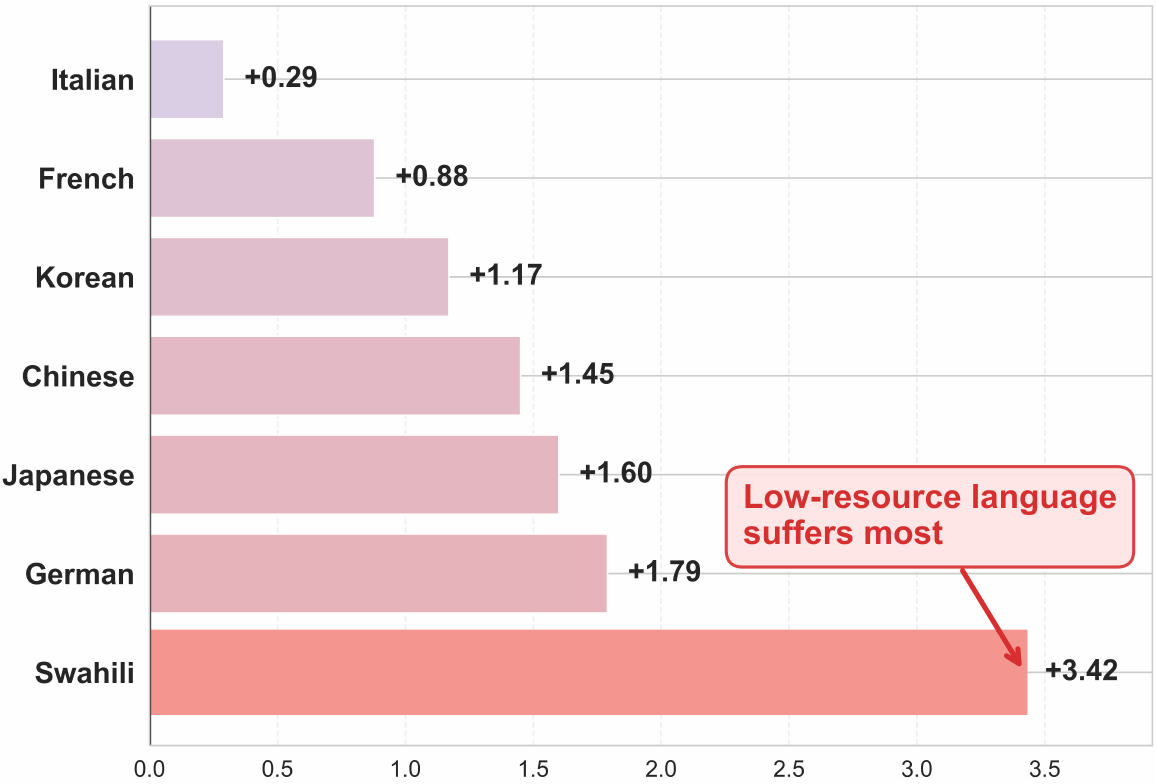}
    \caption{Performance gap between English-thinking and local-language-thinking settings under the same query: average scores of GPT-4o and DeepSeek-3.2 on MMLU-ProX-Health, with the largest degradation in Swahili.}
    \label{fig:performance_gap}
\end{figure}
Medical tasks demand complex reasoning and meticulous deliberation to ensure the safety and reliability of diagnoses~\citep{patel2005thinking, griot2025large}. While reasoning-enhanced Large Language Models (LLMs)~\citep{wei2022chain, jaech2024openai, guo2025deepseek} show significant promise in these life-critical scenarios~\citep{xie2024preliminary}, their capabilities remain uneven across languages. Specifically, models often exhibit substantially stronger reasoning when explicitly prompted to think in English than when prompted to reason directly in the local language~\citep{ranaldi2025multilingual}. As illustrated in the figure~\ref{fig:performance_gap}, this English-as-pivot advantage appears consistently across multiple models, highlighting a persistent multilingual reasoning gap that hinders equitable deployment of medical AI.

Previous efforts to address the multilingual gap follow two main approaches: prompting techniques and cross-lingual post-training. Prompt-based methods~\citep{shi2022language, qi-etal-2025-sot, tam2025language} instruct LLMs to reason in English and then translate outputs to the local language. However, this method introduces systematic limitations: machine translation can be unreliable, especially for low-resource languages~\citep{huang2024evaluating, pang2025salute}, and translation-based reasoning often fails to preserve culturally grounded clinical expertise, leading to factuality misalignment and regional bias~\citep{hu2025large, liu2025translation, schlicht2025disparities, singh2025global}. Cross-lingual training paradigms~\citep{she2024mapo, chai2025xcot, chen-etal-2025-towards-medical} aim to equalize performance via multilingual data exposure, but face complementary challenges: high-quality multilingual medical reasoning data remain scarce and predominantly English-centric~\citep{hu2025large, liu2025translation}, limiting the effectiveness of data-driven approaches.

Both approaches share a common assumption that reasoning must occur primarily in English or in the local language. However, this perspective overlooks a fundamental question: \textbf{what distinct roles might different languages play in medical reasoning?} Recent studies indicate that LLMs perform reasoning in an English-centric way, with key inferential steps shaped heavily by English~\citep{schut2025multilingual, park2025cross}. In contrast, professional medical knowledge is often more accurately preserved in the local language~\citep{hu2025large, liu2025translation}. Building on these findings, we hypothesize a complementary view: pivot-language reasoning provides a transferable logical scaffold (e.g., step-wise structure and consistency checks), whereas local-language reasoning better encodes nuanced, practice-grounded medical knowledge, including region-specific terminology, guideline conventions, and clinically grounded narratives.


Addressing this, we introduce \sysName, a language-informed cross-lingual co-reasoning framework that jointly performs decision-making through parallel English and local-language reasoning. \sysName extracts structured concepts from both chains, uses English as a pivot scaffold, and integrates local clinical signals via concept-level fusion to form a pivot-anchored yet locally grounded reasoning process. It further incorporates retrieval-augmented~\cite{xiong2024benchmarking} to ground the reasoning process in authoritative multilingual medical guidelines. This design aims to improve medical reliability by reducing hallucinations and enhance fidelity by preserving language-specific clinical standards and regional practices.


To comprehensively evaluate multilingual medical reasoning across tasks, we introduce \textbf{\evalName}, a new multilingual benchmark spanning seven non-English languages (Chinese, Japanese, Korean, Thai, Swahili, Zulu, Yoruba) and covering two tasks: long-form question answering and natural language inference. Each instance is annotated by expert physicians, with 350 examples per language. Experiments on \evalName, together with two multiple-choice QA benchmarks (Global-MMLU~\citep{singh2025global} and MMLU-ProX~\citep{xuan2025mmlu}) and extensive ablations, show that \sysName improves both the accuracy and reliability of clinical decision-making, particularly in low-resource language settings. Beyond final-answer correctness, we further assess reasoning quality via automatic proxy evaluation derived from model distillation and expert review, targeting clinical soundness and localization of the generated rationales.

To summarize, our work makes the following novel contributions:
\begin{itemize}
    \item We propose \sysName, leveraging the complementary strengths of English and local-language thinking to focus on reducing reasoning disparities in low-resource languages.
    
    \item We introduce \evalName, a multilingual medical reasoning benchmark covering seven non-English languages and two tasks with special emphasis on three low-resource African languages.
    
    \item We evaluate \sysName across multiple LLM backbones, benchmarks, and tasks in terms of final answer, and further assess reasoning quality using automatic proxy evaluation and expert assessment.
\end{itemize}

\section{Related Work}

\begin{figure*}
    \centering
    \includegraphics[width=\linewidth]{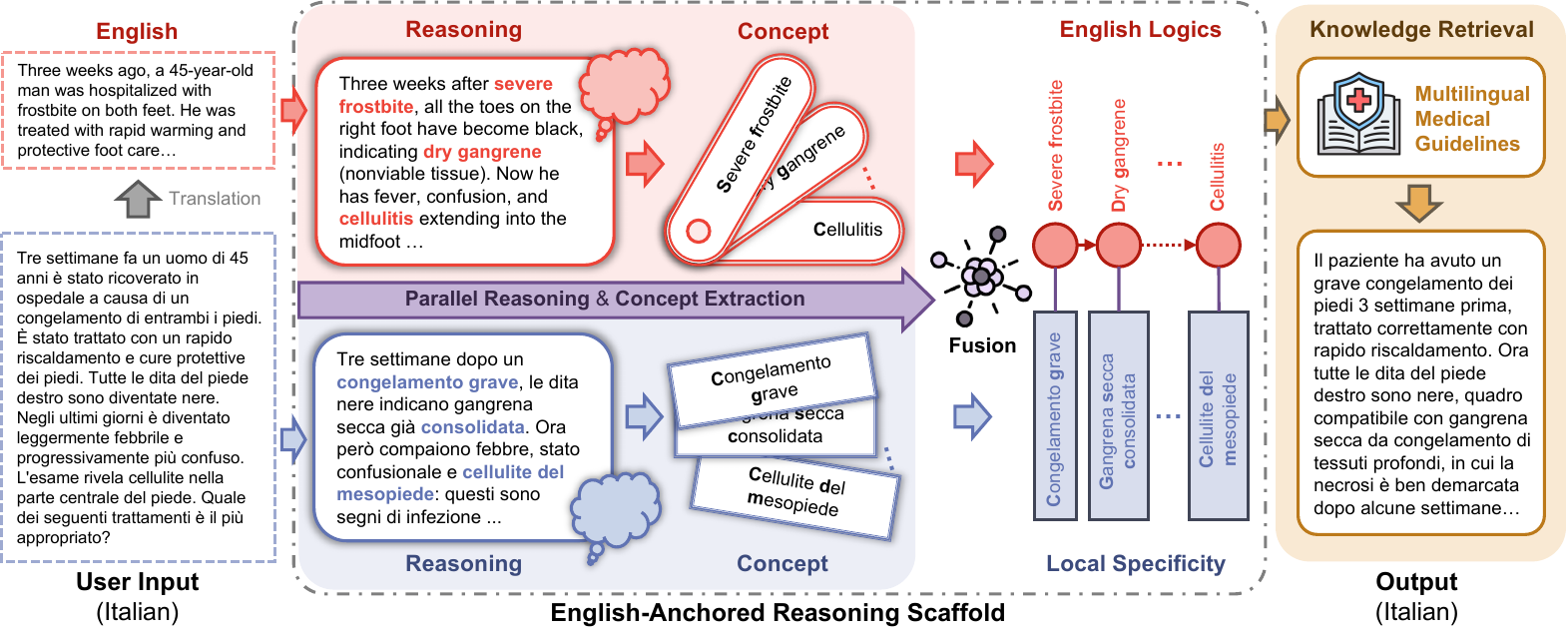}
    \caption{Illustration of the \sysName framework. The framework first translates user input into English, then conducts parallel reasoning in English and Italian via separate queries. Reasoning outputs are abstracted into concepts and fused into an English-anchored reasoning scaffold, where English provides a logical backbone and the local language supplies linguistically specific details. This concept-based scaffold is used to retrieve relevant knowledge and guide the generation of the final Italian reasoning output.}
    \label{fig:framework}
\end{figure*}

\noindent\textbf{Multilingual Medical Reasoning.} Reasoning-centric LLMs such as OpenAI o1~\citep{jaech2024openai} and DeepSeek R1~\citep{guo2025deepseek} leverage test-time computation for step-by-step inference~\citep{wei2022chain}; medical variants further optimize reasoning with verifiable rewards (e.g., HuatuoGPT~\citep{chen-etal-2025-towards-medical}, Med-PRM~\citep{yun2025med}). To address data scarcity, agent pipelines (ReasonMed~\citep{sun2025reasonmed}, MedReason~\citep{wu2025medreason}, MedCaseReasoning~\citep{wu2025medcasereasoning}) synthesize supervision from stronger models, while multi-agent systems (MDAgents~\citep{kim2024mdagents}, MedAgents~\citep{tang2024medagents}) and knowledge-grounded methods~\citep{gao2025txagent, lu2025doctorrag} support complex decisions. Yet research remains pivot-language centric; Qiu et al.~\citep{qiu2024towards} note multilingual needs, but English vs.\ non-English reasoning gaps persist~\citep{shi2022language, kang2025multilingual}. Prior remedies—cross-lingual transfer~\citep{she2024mapo, chai2025xcot}, synthetic data~\citep{singh2024aya, chen2024breaking}, and multilingual CoT~\citep{lu-etal-2024-chain, qi-etal-2025-sot, son2025pushing} often translate English reasoning patterns, leaving open whether reasoning is language-agnostic or English-anchored~\citep{schut2025multilingual, gao2025could}. In this work, we use English as a transferable scaffold while integrating local-language aligned consensus and clinical nuances.

\noindent\textbf{Low-Resource Medical Benchmarks.} Existing work largely centers on single-language medical benchmarks and lacks standardized, parallel evaluation protocols across languages. For instance, IgakuQA~\citep{kasai2023evaluating} targets Japanese, Head-QA focuses on Spanish~\citep{vilares2019head}, FrenchMedMCQA on French~\citep{labrak2022frenchmedmcqa}, RuMedBench on Russian~\citep{blinov2022rumedbench}, while MMedBench~\citep{qiu2024towards} mainly aggregates heterogeneous resources rather than providing a unified parallel benchmark. Available low-resource medical benchmarks are often non-parallel and lack consistent standardization. Moreover, most benchmarks are formulated as multiple-choice question answering (MCQA)~\citep{nimo2025afrimed, singh2025global, xuan2025mmlu}, which limits task diversity and fails to reflect realistic clinical reasoning that requires free-form inference or long-form generation. To fill this gap, we introduce \evalName.


\section{Methodology}
\label{sec:methodology}

In this section, we present \sysName, as illustrated in Fig.~\ref{fig:framework}. We address the fundamental challenge that medical LLMs face significant performance degradation when operating in non-English languages. We first formalize the problem, then detail each component: parallel reasoning generation, cross-lingual concept extraction and fusion, knowledge retrieval, and final answer generation.

\subsection{Problem Formulation}
We formulate multilingual medical question answering as follows. Given a medical $Q_l$ in the local language (e.g., Japanese, Arabic, Swahili, etc), and access to a large language model $\mathcal{M}$ and a multi-lingual medical knowledge base $\mathcal{K}$, our goal is to generate an accurate answer $A_l$ with sound medical reasoning. Formally:
\begin{equation}
    A_l = \mathcal{F}(Q_l, \mathcal{M}, \mathcal{K})
\end{equation}
where $\mathcal{F}$ is our proposed framework.

\subsection{Parallel Reasoning}

Given a question $Q_l$, we generate two independent reasoning paths that capture complementary aspects of medical knowledge: one leveraging the rich English logical thoughts, and another capturing local language contexts. Crucially, these reasoning chains are generated independently without information sharing. This ensures (1) each chain follows its natural reasoning path without bias from the other language; (2) diverse perspectives that can be reconciled through fusion; (3) robustness through redundancy when chains converge on the same conclusion.

\begin{equation}
    R_e = \mathcal{M}(Q_e) ; R_{l} = \mathcal{M}(Q_l)
\end{equation}

where $Q_{e} = \text{translate}(Q_l)$ if the local language is not the English, and we carefully design prompts to encourage step-by-step medical reasoning. 

\subsection{Concept Chain Extraction}

Raw reasoning chains contain verbose natural language that is difficult to align cross-lingually. We extract structured medical concepts to enable precise mapping and fusion. We employ an LLM-based extraction approach that directly identifies medical concepts from reasoning chains.

\begin{equation}
    C = \mathcal{M}(R) = \{c_1\rightarrow c_2\rightarrow \dots \rightarrow c_k\}
\end{equation}

Here, $C$ denotes an ordered concept chain, where $c$ represents a concept and $k$ its index. For each reasoning, the model outputs a list of raw concepts in natural language. As shown in Figure~\ref{fig:framework}, English reasoning is abstracted as $C_e=\{\text{Severe frostbite}\rightarrow \text{Dry gangrene}\rightarrow \dots \rightarrow \text{Cellulitis}\}$, while Italian reasoning is extracted as $C_l=\{\text{Congelamento grave}\rightarrow  \dots \rightarrow \text{Cellulite mesopiede}\}$. The ordering preserves the logical coherence in the reasoning process.


\subsection{Cross-lingual Concept Fusion}

The concept fusion module integrates English and local language concept chains into a unified representation, enabling us to leverage the strengths of both languages while maintaining logical consistency and semantic coherence. 

\paragraph{Fusion Strategy.} Algorithm~\ref{alg:concept_fusion} in Appendix~\ref{sec:full_fusion_strategy} details our position-aware backbone-augmentation fusion strategy. In summary, we treat the English concept chain $C_e$ as the backbone and augment it with local-language concepts that provide complementary clinical information. Specifically, we initialize $C_f \leftarrow C_e$, then for each $c_l \in C_l$ we compute its maximum embedding similarity to concepts in $C_e$; if the score exceeds a threshold $\tau$, we add $c_l$ to $C_f$, anchored to its most similar English concept. We adopt BGE-M3~\citep{chen2024bge} as the multilingual embedding model and set $\tau$ to $0.5$. We get an English-anchored reasoning scaffold by: 
\begin{equation}
    C_{\text{f}} = C_e \cup \{\, c_l \in C_l \mid \max_{c_e \in C_e} \text{sim}(c_l, c_e) > \tau \,\}
\end{equation}

This design is motivated by: (1) Logicality, leveraging the superior consistency of multi-step English reasoning; (2) Complementarity, integrating culture-specific medical knowledge embedded in local language; (3) Conceptual Alignment, ensuring that key medical concepts are faithfully addressed across linguistic contexts.



\subsection{Final Answer Generation} 

\noindent\textbf{Knowledge Retrieval.} 
The fused concept chain $C_{f}$ serves as the structural backbone of the reasoning process. However, as $C_{f}$ represents highly compressed information, it functions primarily as a reasoning root that requires further elaboration to ensure clinical utility. Moreover, to enhance medical reliability while preserving language-specific clinical standards and regional practices, we introduce a knowledge-enrichment phase that expands these abstract nodes with verifiable, evidence-based information. Specifically, to account for regional heterogeneity in medical knowledge and clinical guidelines, we construct a multilingual knowledge base derived from the MSD Manuals.~\citep{msdmanuals-professional}, integrated with official permission. For low-resource African languages, we additionally incorporate medical materials from AFRIDOC-MT~\citep{alabi-etal-2025-afridoc}. Specifically, we use questions both in English as well as the local language to retrieve top-3 relevant documents $D$ via the BGE-M3 retriever from the corresponding language-specific knowledge base. This grounding strategy ensures that reasoning is supported by evidence aligned with regional and linguistic contexts. More implementation details can be found in Appendix~\ref{sec:detailed_implementation}.

\noindent\textbf{Answer Generation.}
Guided by the original query $Q_l$, the fused concept chain $C_{f}$, and the retrieved evidence $D$, the model is prompted to synthesize a response. In this stage, $C_{f}$ serves as the structural reasoning trajectory, while the retrieved documents $D$ provide the necessary empirical grounding. By aligning the abstract logic of the concept chain with the concrete clinical data, the model generates a final, verifiable response in the target language: $A_l = M(A_l, C_{f}, D)$.

%


\section{Experiment}
\begin{table*}[t]
\centering
\scriptsize
\setlength{\tabcolsep}{2.8 pt}
\renewcommand{\arraystretch}{1.3}

\begin{tabular}{lcccccccccccccccccc}
\toprule
\multirow{2}{*}{\textbf{Method}} & \multicolumn{9}{c}{\textbf{Global-MMLU-Medical}} & \multicolumn{9}{c}{\textbf{MMLU-ProX-Health}}  \\
\cmidrule(lr){2-10}  \cmidrule(lr){11-19}
 & \textbf{ZH} & \textbf{JA} & \textbf{KO} & \textbf{DE} & \textbf{FR} & \textbf{ES} & \textbf{IT} & \textbf{SW} & \textbf{Avg.}  & \textbf{ZH} & \textbf{JA} & \textbf{KO} & \textbf{DE} & \textbf{FR} & \textbf{ES} & \textbf{IT} & \textbf{SW} & \textbf{Avg.} \\
\midrule

\multicolumn{19}{l}{\textit{Closed-Source Models}} \\
\midrule
\midrule

Claude-3.5-haiku  & 64.78 & 66.32 & 64.25 & 72.55 & 72.73 & 75.80 & 72.58 & 56.65 & 68.21  & 56.33 & 57.64 & 54.88 & 60.70 & 61.72 & 59.97 & 62.59 & 35.81 & 56.21 \\

\rowcolor{theme_purple!50}GPT-4o  & 82.39 & 82.66 & 81.45 & 82.59 & 83.26 & 83.39 & 82.46 & 76.08 & 81.79  & 67.39 & 67.39 & 65.94 & 69.14 & 69.72 & 70.60 & 71.47 & 61.86 & 67.94 \\

\rowcolor{theme_red!40}GPT-5.1  & 83.72 & 84.18 & 82.86 & 85.18 & 86.30 & 85.71 & 85.77 & 79.19 & 84.11  & 71.32 & 70.45 & 70.45 & 71.32 & 72.78 & 72.63 & 74.24 & 67.89 & 71.39\\

GPT-5.2  & 84.39 & 86.25 & 83.65 & 85.18 & 86.25 & 85.98 & 85.78 & 81.86 & 84.92  & 72.63 & 73.94 & 73.94 & 76.42 & 77.00 & 75.69 & 75.98 & 70.60 & 74.53\\
\midrule 
\rowcolor{theme_purple!50}CoT  & 81.06 & 75.42 & 77.74 & 81.93 & 84.25 & 83.32 & 81.93 & 74.42 & 80.00  & 67.10 & 61.28 & 63.61& 70.31 & 72.34 & 70.01 & 71.32 & 60.99 & 67.12 \\
\rowcolor{theme_purple!50}SoT  & 81.59 & 80.53 & 78.34 & 81.53 & 82.46 & 82. 59 & 81.73 & 74.55 & 80.42  & 65.21 & 64.63 & 62.45 & 67.69 & 68.85 & 68.27 & 69.72 & 56.19 & 65.37 \\
\rowcolor{theme_purple!50}Self-Consistency  & 83.77 & 82.79 & 81.61 & 83.32 & 84.29 & 84.24 & 82.62 &  77.31 & 82.49  & 68.08 & 68.51 & 66.96 & 70.12 & 71.18 & 71.72 & 71.97 & 62.39 & 68.87 \\
\rowcolor{theme_purple!50}RAG + CoT  & 82.92 & 81.86 & 81.26 & 83.46 & 83.52 & 84.05 & 83.39 & 77.94 & 82.30  &69.14 & 69.87 & 67.83 & 70.31 & 72.63 & 70.89 & 72.63 & 65.60 & 69.86\\

\rowcolor{theme_purple!50}Ours (GPT-4o)  & 84.98 & 83.78 & 83.77 & 84.85 & 85.38 & 84.91 & 84.45 & 81.52 & 84.20  & 71.90 & 71.32 & 71.62 & 71.76 & 73.22 & 73.21 & 73.07 & 69.14 & 71.91 \\

\rowcolor{theme_red!40} Ours (GPT-5.1)   & \textbf{89.10} & \textbf{88.02} & \textbf{88.90} & \textbf{88.09} & \textbf{88.47} & \textbf{89.54} & \textbf{89.16} & \textbf{87.62} & \textbf{88.61}   & \textbf{76.56} & \textbf{77.87} & \textbf{76.85} & \textbf{77.72} & \textbf{77.43} & 
\textbf{78.60} & \textbf{78.31} & \textbf{75.98} & \textbf{77.42} \\

\midrule
\multicolumn{19}{l}{\textit{Open-Source Models}} \\
\midrule
\midrule
\rowcolor{theme_orange!60}DeepSeek-3.2  & 81.06 & 80.33 & 77.54 & 81.59 & 83.19 & 83.59 & 81.33 & 68.11 & 79.59 & 66.81 & 66.81 & 61.28 & 69.00 & 68.70 & 67.89 & 69.00 & 51.67 & 65.15\\

\rowcolor{theme_green!60}Qwen3-30B  & 77.54 & 76.21 & 72.69 & 77.87 & 79.40 & 78.67 & 77.48 & 51.03 & 73.86 & 61.86 & 56.91 & 54.15 & 61.72 & 61.72 & 64.48 & 63.32 & 27.22 & 56.42 \\

Qwen2.5-72B  & 79.34 & 77.81 & 72.96 & 76.61 & 79.93 & 80.07 & 78.60 & 50.90 & 74.52 & 59.39 & 56.48 & 54.15 & 59.83 & 61.86 & 61.51 & 60.26 & 30.13 & 55.45  \\
LLaMA3.1-70B  & 73.02 & 72.36 & 67.38 & 76.08 & 77.94 & 78.54 & 78.14 & 65.58 & 73.63 & 51.09 & 48.33 & 47.45 & 59.53 & 60.84 & 60.26 & 60.84 & 44.25 & 54.07 \\

Qwen2.5-32B  & 77.21 & 73.95 & 68.17 & 75.68 & 76.68 & 76.61 & 75.08 & 51.76 & 71.89 & 58.22 & 54.73 & 48.33 & 59.10 & 58.66 & 60.41 & 58.95 & 26.76 & 53.15\\

\midrule
\rowcolor{theme_green!60}Ours (Qwen3-30B)  & 80.23 & 78.80 & 75.48 & 79.67 & 80.27 & 80.60 & 79.14 & 50.76 & 75.62 & 64.92 & 62.15 & 59.39 & 65.65 & 66.38 & 67.83 & 68.41 & 39.01 & 61.72\\
\rowcolor{theme_orange!60}Ours (DeepSeek-3.2)  & \textbf{85.85} & \textbf{83.39} & \textbf{86.58} & \textbf{85.45} & \textbf{86.17} & \textbf{86.57} & \textbf{85.31} & \textbf{82.19} & \textbf{85.19} & \textbf{71.91} & \textbf{71.91} & \textbf{72.34} & \textbf{73.36} & \textbf{74.52} & \textbf{73.22} & \textbf{73.91} & \textbf{68.85} & \textbf{72.50}\\
\bottomrule
\end{tabular}
\caption{Results on Global-MMLU and MMLU-ProX.
CoT, SoT, and Self-Consistency are reasoning strategies, with SoT specifically enhancing multilingual reasoning. The highest performance scores are shown in \textbf{bold}. Different model backbones are distinguished using background colors:
\sethlcolor{theme_purple!50}\hl{GPT-4o},
\sethlcolor{theme_red!40}\hl{GPT-5.1},
\sethlcolor{theme_green!60}\hl{Qwen3-30B}, and
\sethlcolor{theme_orange!60}\hl{DeepSeek-v3.2}.}
\label{tab:mcqa-result}
\end{table*}

\subsection{Evaluation Benchmark}

\begin{figure*}
    \centering
    \includegraphics[width=\linewidth]{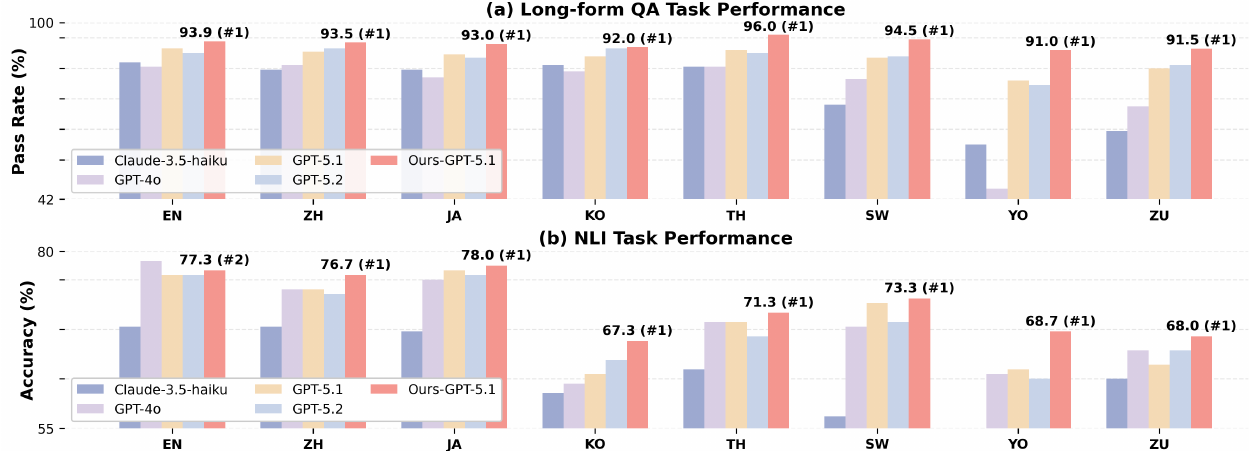}
    \caption{Experimental results on \evalName, where (\#) denotes the ranking of our framework.}
    \label{fig:multimed-bar}
\end{figure*}

\noindent\textbf{Global-MMLU and MMLU-ProX.} To evaluate multilingual medical reasoning, we use the medical subsets of two major benchmarks: Global-MMLU~\citep{singh2025global}, which emphasizes linguistic and cultural diversity, and MMLU-ProX~\cite{xuan2025mmlu}, which targets challenging cross-linguistic reasoning. Specifically, we select the medical subset of Global-MMLU and the health category of MMLU-ProX, with 1,505 and 687 items per language, respectively. We evaluate the following languages: Western languages - German (DE), French (FR), Spanish (ES), and Italian (IT); Asian languages - Chinese (ZH), Japanese (JA), and Korean (KO); and an African language - Swahili (SW). Both benchmarks are structured as multiple-choice question-answering (MCQA).

\noindent\textbf{\evalName.} To evaluate the proposed method beyond MCQA, we introduce \evalName, a multilingual medical benchmark including two additional task settings: natural language inference (NLI) and open-ended long-form question-answering (LFQA), covering seven non-English languages. We sample 150 instances from the English BioNLI \cite{bastan-etal-2022-bionli} dataset and 200 instances from LiveQA \cite{qianying-etal-2020-liveqa}, and construct multilingual versions via machine translation. Each translated instance is independently reviewed and revised by two native bilingual experts for each target language, except for Yoruba, which is reviewed by one expert. The expert team comprises approximately 12 physicians or senior medical students, providing domain knowledge to support the accuracy and consistency of the annotations. The resulting \evalName spans seven non-English languages: ZH, JA, KO, Swahili (SW), Thai (TH), Yoruba (YO), and Zulu (ZU).

\subsection{Experimental Setups}

\noindent\textbf{Evaluation Metrics.} We evaluate each task as follows: For the MCQA and NLI tasks, we report accuracy based on an exact match criterion. For the LFQA task, we employ GPT-4o as an automated judge~\cite{li2025generation} to score responses on a 5-point Likert scale across five dimensions: \textit{Overall Quality}, \textit{Correctness}, \textit{Completeness}, \textit{Safety}, and \textit{Hallucination}. The complete evaluation introduction is shown in Appendix~\ref{sec:detailed_implementation}. Additionally, we calculate a pass rate, defined as the percentage of responses where both the Overall Quality and Safety scores are 4 or higher.

\noindent\textbf{Baselines.} We evaluate multilingual medical reasoning across both closed- and open-source models. Closed-source models include Claude-3.5-Haiku~\citep{anthropic2024_3-5_models_computer_use} and the GPT family (GPT-4o, GPT-5.1, and GPT-5.2)~\citep{hurst2024gpt, openai2025_gpt5.1}. Open-source models include DeepSeek-3.2~\citep{liu2025deepseek}, LLaMA3.1-70B-Instruct~\citep{grattafiori2024llama}, and Qwen instruction models (Qwen2.5-72B/32B and Qwen3-30B)~\citep{yang2025qwen3}). We also compare multiple reasoning strategies: Chain-of-Thought (CoT)~\citep{wei2022chain}, Structured-of-Thought (SoT)~\citep{qi-etal-2025-sot}, Self-consistency~\citep{wang2022self}, and a vanilla CoT-enhanced RAG pipeline using our custom knowledge base.

\noindent\textbf{Implementation Details.} We evaluate \sysName with four backbones: GPT-4o, GPT-5.1, Qwen3-30B-Instruct and DeepSeek-3.2. All closed-source models, as well as DeepSeek-3.2, are accessed via APIs, while the remaining models are run locally on a cluster of eight 40GB A100 GPUs. We consistently set a sampling temperature of 0.7 and apply a low reasoning effort to the reasoning models.

\subsection{Main Results}

Table~\ref{tab:mcqa-result} presents overall results on Global-MMLU and MMLU-ProX. Figures~\ref{fig:multimed-bar} shows comparative results on \evalName, including LFQA pass rates and NLI accuracy; Figure~\ref{fig:multimed-radar} details the dimensional scores for LFQA. Full score statistics are provided in Appendix~\ref{sec:full-results}. Based on this analysis, we draw the following conclusions:

\noindent\textbf{Superior performance across multiple evaluation paradigms.}
\sysName demonstrates robust improvements across both MCQA and LFQA tasks. On MCQA benchmarks, the \sysName on GPT-5.1 backbone shows substantial gains, with an average improvement of $4.5$ points on Global-MMLU and $6.03$ points on MMLU-Pro. Notably, it consistently outperforms established reasoning baselines, indicating that our cross-lingual reasoning architecture provides synergistic benefits. On the \evalName LFQA benchmark, \sysName achieves complementary gains in response quality, attaining the highest overall scores across all eight languages, with particularly notable improvements in \textit{completeness}. These advancements across diverse tasks validate that \sysName enhances multiple cognitive processes, including coherent medical reasoning and comprehensive information synthesis.

\noindent\textbf{Larger benefits for low-resource languages.} A critical finding is that \sysName provides disproportionately larger improvements for underrepresented languages, directly addressing performance gaps. For low-resource African languages, our framework achieves remarkable gains: Swahili improves by over 8 points on both Global-MMLU and MMLU-Pro, while Yoruba shows a $+9.0\%$ increase in pass rate on LFQA. This pattern suggests our method effectively compensates for the base model's weaker reasoning capabilities in low-resource settings. By maintaining a parallel reasoning strategy, \sysName enables models to leverage superior medical reasoning in English while preserving culturally specific clinical nuances. The resulting convergence of performance across languages demonstrates a substantial reduction in linguistic disparity for medical tasks.

\noindent\textbf{Enhanced reasoning depth and safety without accuracy trade-offs.} \sysName achieves superior response quality and comprehensiveness while maintaining strong factual accuracy. On \evalName, our framework shows substantial improvements in completeness scores and reduced hallucination rates across all languages, while maintaining competitive NLI accuracy. This indicates that \sysName excels at producing comprehensive, clinically sound responses rather than merely optimizing for surface-level correctness, effectively balancing reasoning depth with precision.
\begin{figure}
    \centering
    \includegraphics[width=0.8\linewidth]{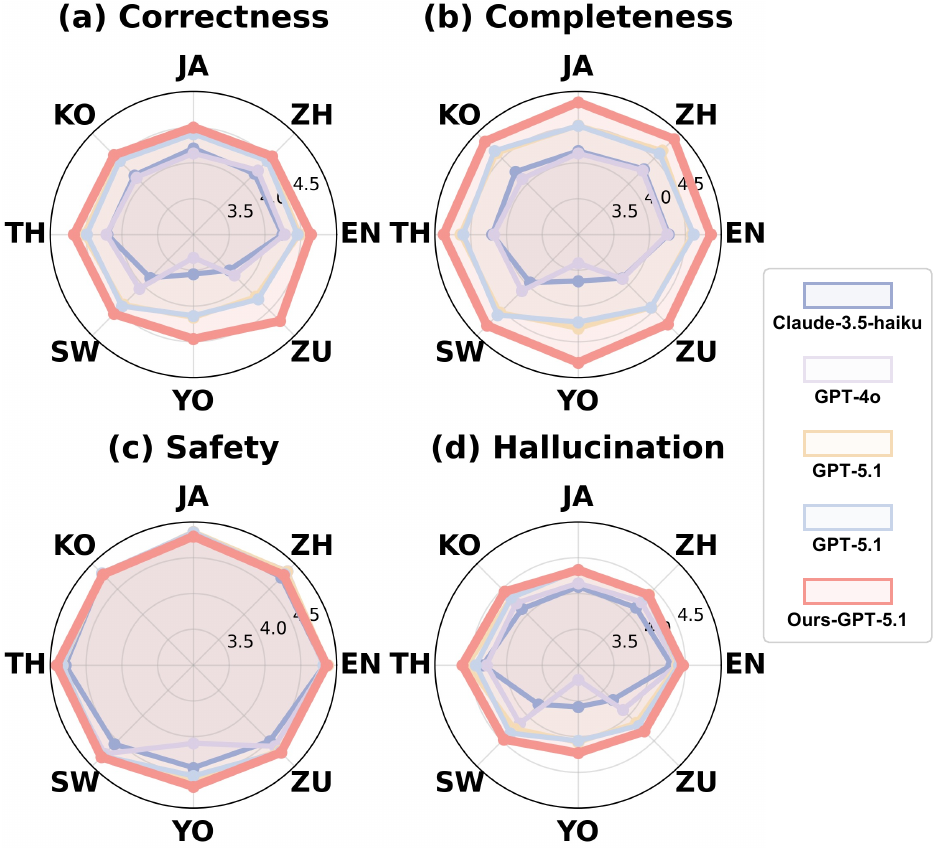}
    \caption{Results on LFQA, judged by GPT-4o.}
    \label{fig:multimed-radar}
\end{figure}

\section{Ablation Study}
To understand the contribution of each component in \sysName, we conduct an ablation study by removing individual models and measuring the impact across benchmarks and languages.

\noindent\textbf{Configuration.} We evaluate three configurations: (1) W/O RAG: no knowledge retrieval; (2) W/O English: remove the English reasoning chain and English RAG; (3) W/O local: remove the local-language reasoning chain and local-language RAG. We run experiments on MMLU-ProX and the LFQA task using three African languages.
\begin{table*}
    \centering
    \scriptsize
    \setlength{\tabcolsep}{5pt}
    \begin{tabular}{llcccccccc}
        \toprule
        \textbf{Backbone} & \textbf{Train Set}& \textbf{Chinese} & \textbf{English} & \textbf{French} & \textbf{Japanese} & \textbf{Russian} & \textbf{Spanish} & \textbf{Avg.} \\
        \midrule
        \multirow{2}{*}{Gemma-7B-it}& MMedBench & 56.07 & 52.16 & 34.89 & 34.67 & 63.28 & 57.51 & 49.45 \\
        & \dataName & 52.38(\textcolor{red}{-3.69}) & 52.08(\textcolor{red}{-0.08}) & 40.03(\textcolor{ForestGreen}{+5.14}) & 41.71(\textcolor{ForestGreen}{+7.04}) & 64.45(\textcolor{ForestGreen}{+0.55}) & 58.06(\textcolor{ForestGreen}{+3.48}) & 51.39(\textcolor{ForestGreen}{+1.94}) \\
        \midrule
        \multirow{2}{*}{Qwen2.5-7B}& MMedBench & 81.47 &61.67 & 47.72 & 48.74 & 69.53 & 67.18 & 62.16 \\
        & \dataName & 78.78(\textcolor{red}{-3.18})&62.37(\textcolor{ForestGreen}{+0.70}) & 54.66(\textcolor{ForestGreen}{+6.94}) & 56.78(\textcolor{ForestGreen}{+8.04}) & 70.31(\textcolor{ForestGreen}{+0.78}) & 69.22(\textcolor{ForestGreen}{+2.04}) & 64.98(\textcolor{ForestGreen}{+2.82}) \\
        \midrule
        \multirow{2}{*}{Qwen2.5-14B} &MMedBench &84.47 &71.48 & 64.15 & 66.33 & 75.39 & 78.77 & 73.11 \\
        & \dataName & 82.89(\textcolor{red}{-1.58})& 75.73(\textcolor{ForestGreen}{+4.25})&72.03(\textcolor{ForestGreen}{+7.88}) & 68.34(\textcolor{ForestGreen}{+2.01}) & 75.78(\textcolor{ForestGreen}{+0.39}) & 82.39(\textcolor{ForestGreen}{+3.62}) & 75.97(\textcolor{ForestGreen}{+2.86})\\
        \bottomrule
    \end{tabular}
    \caption{Impact of training data on cross-lingual performance: comparison across languages on MMedBench.}
    \label{tab:mmedbench-result}
\end{table*}

\begin{figure}
    \centering
    \includegraphics[width=\linewidth]{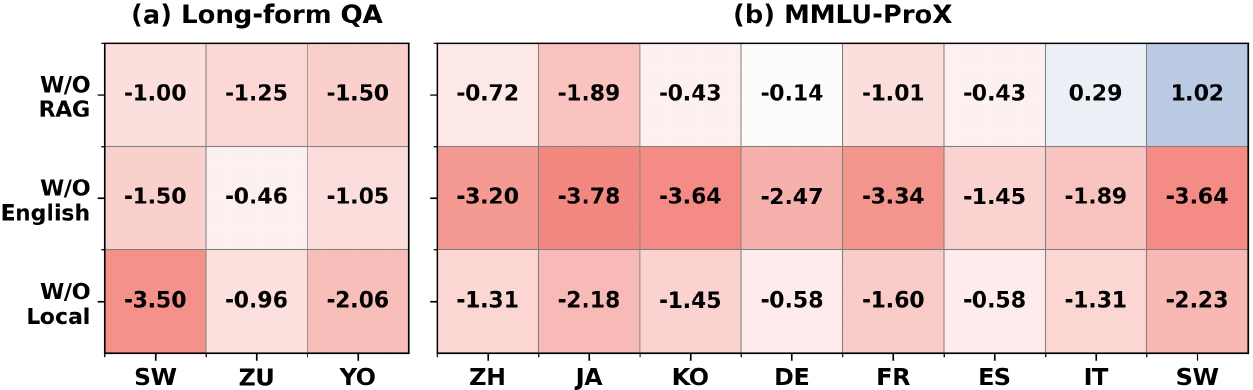}
    \caption{Ablation results on selected languages across LFQA and MMLU-ProX.}
    \label{fig:ablation-study}
\end{figure}
Results are shown in Figure~\ref{fig:ablation-study}. We summarize the key findings:
\noindent\textbf{(1) The utility of each reasoning language depends on the task.} For the complex reasoning in MMLU-ProX, English reasoning provides a strong scaffold, especially for lower-resource languages. Conversely, for the culturally-grounded long-form QA task, local-language reasoning is critical—its removal causes the largest performance drops, particularly in Swahili and Yoruba. This supports our hypothesis that while English supplies a robust reasoning framework, local-language reasoning preserves culturally-specific nuances.
\noindent\textbf{(2) The importance of local-language reasoning increases for lower-resource settings.} While high-resource languages (e.g., Chinese, German) experience moderate degradation without local-language reasoning, the impact is more severe for lower-resource languages. This pattern is amplified in the LFQA task, where ablation leads to absolute performance losses between 1.0 and 3.5 points. This indicates that local language captures essential, culturally-grounded concepts and terminology that English alone cannot fully represent in low-resource languages.
\noindent\textbf{(3) RAG provides consistent but variable gains, with quality-dependent exceptions.} While knowledge retrieval generally improves performance, its impact is moderate and uneven across languages. Notably, Italian and Swahili exhibit slight performance declines when RAG is used, suggesting retrieved documents can sometimes introduce noise or contradictions. This highlights the limitations of our simple RAG techniques, especially for low-resource languages, pointing to the need for future work on improved relevance filtering and reliable source retrieval in the medical domain.

    

\begin{table}
    \centering
    \small
    \setlength{\tabcolsep}{1.2pt}
    \begin{tabular}{ccccc}
    \toprule
     Ours vs. GPT-5.1   & \textbf{Clarity} & \textbf{Soundness} & \textbf{Safety} & \textbf{Localization} \\
    \midrule
        Win Rate (\%) & 52.5 & 45.0 & 40.0 & 52.5 \\ 
        Tie Rate (\%) & 22.5 & 27.5 & 50.0 & 32.5 \\
    \bottomrule
    \end{tabular}
    \caption{Results of pairwise comparison by native physicians. Detailed examples are shown in Appendix~\ref{sec:expert_evaluation}. }
    \label{tab:expert-results}
\end{table}
\section{Quality and Robustness Analysis}

To evaluate the quality, robustness, and reliability of \sysName's generated reasoning, we conduct three complementary evaluations: automated comparison through model distillation, expert human assessment, and attack-style analysis.

\subsection{Model Distillation}

A key challenge in evaluating medical reasoning is that standard metrics assess only final answers, not the reasoning process. To address this, we use \textit{model distillation} as a proxy: if a reasoning chain embodies valid medical knowledge and logic, a model trained on it should perform better~\citep{hinton2015distilling, xu2024survey}. We use \sysName to construct reasoning training data and evaluate its effectiveness by fine-tuning models and testing their performance on medical reasoning tasks with MMedBench~\citep{qiu2024towards}.


\noindent\textbf{Data Construction.} Our source data is the MMedBench training set, which contains medical questions and corresponding rationales in six languages. However, these original rationales have a critical limitation: they are \textit{retrospective explanations} authored with knowledge of the correct answer, rather than reflecting a forward, step-by-step clinical reasoning process. Such post-hoc rationales often lack the uncertainty and differential decision-making inherent to real-world practice~\citep{zuo2025medxpertqa}. To address this, we apply \sysName with GPT-5-mini to generate forward reasoning traces. We sample 10,000 questions each from the Chinese and English subsets and use all available data for the remaining languages. For each question, \sysName produces a reasoning chain, with the final response used as the new rationale. We retain only items with correct answers, forming our new dataset \dataName. Full statistics are provided in Appendix~\ref{sec:training_data}.

\noindent\textbf{Implementation.} We fine-tune three models of varying capabilities: Gemma-7B-it~\citep{team2024gemma}, Qwen-2.5-7B-Instruct and Qwen3-14B, using both the original MMedBench training data and our newly constructed \dataName. Fine-tuning is performed with LoRA~\citep{hu2022lora} (rank 8), a learning rate of 1.0e-4, over 3 epochs.

\noindent\textbf{Results.} Comprehensive results are provided in the Table~\ref{tab:mmedbench-result}. While the performance on Chinese questions shows a slight decrease due to our sampling strategy, we observe significant and consistent improvements when models are trained on \dataName compared to the original MMedBench data, particularly on tasks requiring complex reasoning. For example, the French subset includes questions with single or multiple correct answers, a format that demands careful logical discrimination. On this subset, \dataName achieves an improvement of over 5 points across all model backbones. These gains, consistent across multiple languages, demonstrate the high quality and generalizability of the reasoning processes captured in \dataName.

\begin{figure*}[ht]
    \centering
    \includegraphics[width=\linewidth]{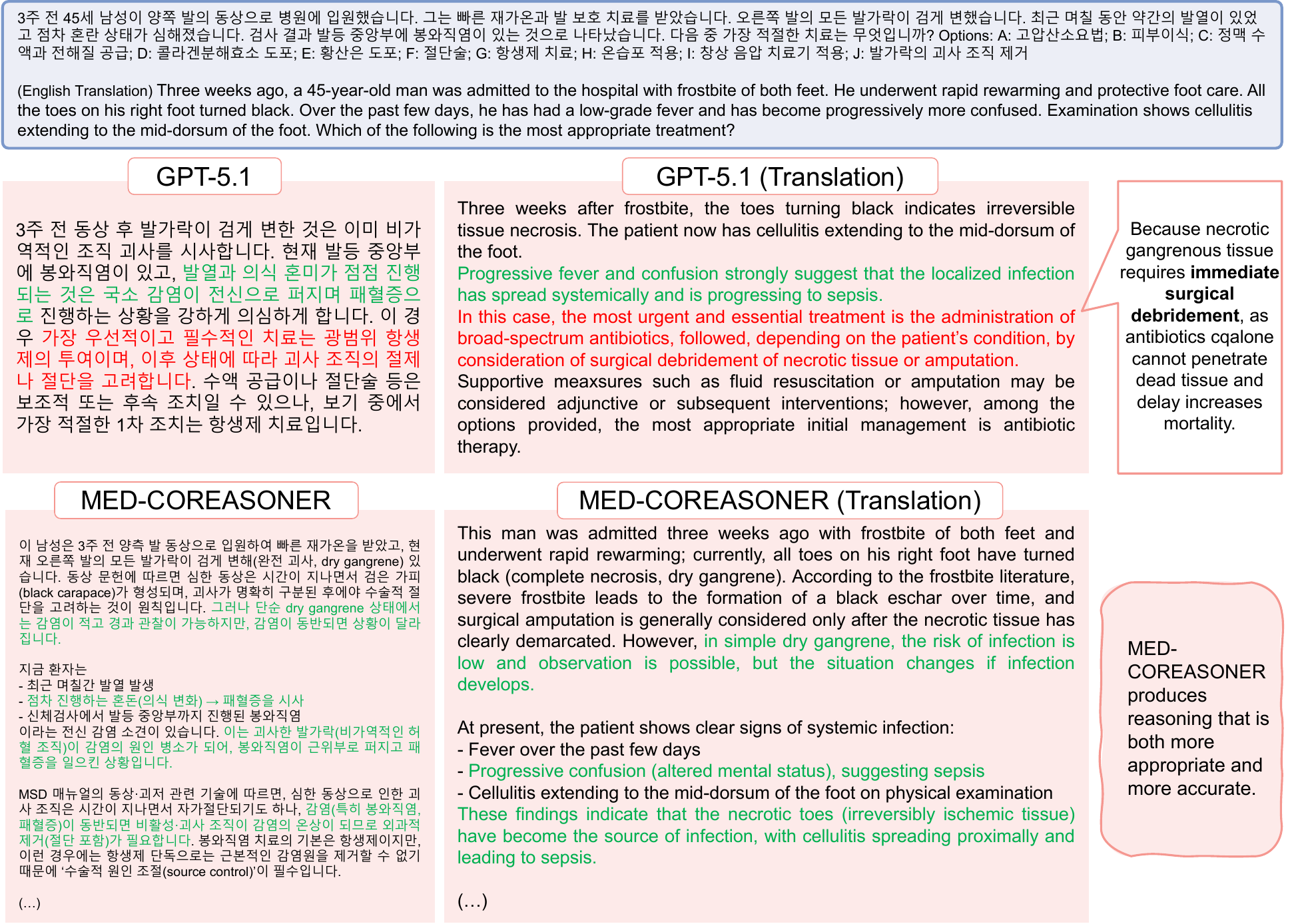}
    \caption{Korean example of correctness and incorrectness annotation in the expert evaluation. \textcolor{ForestGreen}{Green} marks sentences or phrases judged to be clinically correct and appropriate. \textcolor{red}{Red} marks sentences or phrases judged to be clinically incorrect, unsupported, potentially unsafe, or misleading.}
    \label{fig:case_study}
\end{figure*}

\subsection{Expert Clinical Evaluation}
To evaluate nuanced clinical reasoning beyond automated metrics, we conduct a blinded expert study with four native-speaking physicians (Spanish, Chinese, Korean, and Japanese). The study consists of two complementary evaluation tasks over anonymized responses from \sysName and GPT-5.1 on clinical questions drawn from MMLU-ProX. In the first task, each physician performs pairwise comparisons on 10 randomly sampled question answer pairs in their language. To ensure a fair comparison of reasoning quality rather than final answer accuracy, we include only questions for which both models produce the correct final answer, resulting in 40 question answer pairs in total across the four languages. For each pair, experts assess the responses along four dimensions: \textit{Clarity}, which captures logical flow and coherence of the explanation; \textit{Soundness}, which measures medical accuracy and clinical appropriateness; \textit{Safety}, which evaluates whether the response avoids potentially harmful or misleading recommendations; and \textit{Localization}, which reflects how well the response aligns with regional medical practice, conventions, and terminology. In the second task, each physician is additionally shown one clinical question together with two anonymized candidate answers from \sysName and GPT-5.1, and is asked to annotate the reasoning in each response by identifying correct and incorrect clinical justifications. The full guidelines are provided in the Appendix~\ref{sec:expert_evaluation}.



As shown in Table~\ref{tab:expert-results}, \sysName demonstrates strong performance, particularly in Localization and Clarity. This validates that its explicit parallel reasoning produces culturally-grounded and well-structured outputs. While both models show competitive Clinical Soundness, \sysName achieves a $90.0\%$ win+tie rate on Safety, indicating reliably safer generations. Fig.~\ref{fig:case_study} presents a Korean case study illustrating the labeling of correctness and incorrectness in the reasoning. Overall, these results confirm that \sysName achieves comparable clinical quality while offering superior reasoning clarity and effective cross-lingual knowledge transfer. 

\subsection{Attack-Style Robustness Analysis}

To examine the robustness of \sysName, we conduct an attack-style analysis by injecting incorrect concept alignments during concept fusion. In this analysis, we use French as the test language and simulate concept poisoning by randomly selecting $n$ local concepts and mis-anchoring each to the $K$-th most similar English concept rather than the top match, thereby introducing plausible but incorrect alignments into the fused concept chain. We set $K{=}5$, keep all other settings unchanged, and report results on MMLU-ProX to measure the sensitivity of downstream reasoning and answer quality to such errors.



\begin{table}[]
    \centering
    \begin{tabular}{cccccc}
    \toprule
     no attack & $n{=}1$ & $n{=}3$ & $n{=}5$     &   $n{=}10$\\
    \midrule
    77.43 &	76.42&76.71 &	76.56&75.54 \\
    \bottomrule

    \end{tabular}
    \caption{Accuracy in attack-style analysis of \sysName, where $n$ local concepts are deliberately mis-anchored during concept fusion stage.}
    \label{tab:attack}
\end{table}
Table~\ref{tab:attack} shows that performance remains largely stable when a small number of incorrect concepts are injected ($n{=}1,3,5$), with marginal declines relative to the no-attack setting. This suggests that \sysName is robust to moderate levels of hallucinated concept insertion in the fused chain.

\section{Conclusion}

In this work, we explore the reasoning gap between English-centric and local-language thinking in medical contexts. We propose \sysName, a framework that combines the logical rigor of English with the clinical specificity of local-language reasoning. To evaluate multilingual medical reasoning, we introduce \evalName, covering diverse tasks with an emphasis on low-resource languages. Experiments on three benchmarks show that \sysName improves medical reasoning accuracy. Ablation studies further reveal that local-language reasoning is especially beneficial in low-resource settings. Evaluation via model distillation and expert review confirms that \sysName enhances both reasoning clarity and local clinical relevance.

\section*{Limitations}
While \sysName demonstrates strong performance across benchmarks, it has several limitations:
(1) \textit{Unexplored theoretical grounding}: Our experiments and ablation studies show that removing English reasoning leads to significant performance drops, particularly on complex reasoning tasks, suggesting that English plays a critical role in providing logical structure. However, we do not offer a theoretical analysis of how different language modes contribute to reasoning.
(2) \textit{Dependency on English as pivot}: We currently use English as the sole pivot language for reasoning. The potential of other pivot languages (e.g., Chinese) remains unexplored and may offer complementary benefits.
(3) \textit{Computational overhead and efficiency considerations}. \sysName adopts a multi-stage architecture that enables parallel generation of dual reasoning chains but requires sequential concept extraction, fusion, knowledge retrieval, and synthesis, resulting in higher API usage and latency than single-pass approaches. Despite this additional overhead, \sysName delivers substantial performance gains, particularly in low-resource languages such as Swahili and Yoruba, demonstrating clinically meaningful improvements in accuracy, completeness, and safety. The cost-benefit trade-off is favorable for non-urgent clinical applications where decision quality is prioritized over response speed. Moreover, some optimization strategies could mitigate computational costs without sacrificing performance: (a) implementing an adaptive RAG that triggers only for complex queries; (b) distilling the multi-stage reasoning into more efficient student models.
(4) \textit{Expert Annotation in Yoruba}. We include only one expert to review the Yoruba translations during the construction of Multimed-X, and we will continue refining this subset in future work.

\section*{Ethical Considerations}
All data used in this paper comply with privacy and licensing requirements. The medical knowledge base corpus is constructed from the MSD Manuals with official permission. All other datasets are obtained from publicly available open-source repositories. Expert annotators for \evalName and physicians involved in assessment experiments are formally recruited and compensated or included as co-authors on the paper.

\section*{Acknowledgments}
Dr. Irene Li is supported by JST ACT-X (Grant JPMJAX24CU) and JSPS KAKENHI (Grant 24K20832). This work used supercomputers provided by the Research Institute for Information Technology, Kyushu University, through the HPCI System Research Project (Project ID: \texttt{hp250092}). This work is also supported by Google Academic Research Award 2025. Dr. Hyunjae Kim and Dr. Qingyu Chen are supported by the National Institutes of Health National Library of Medicine under Award Number R01LM014604. We sincerely thank Dr. Yong Hoe Koo (Asan Medical Center, University of Ulsan College of Medicine, South Korea) for his expert contribution to the clinical evaluation study. 
\bibliography{custom}
\clearpage
\appendix
\section{Position-aware Concept Fusion Strategy}
\label{sec:full_fusion_strategy}
Algorithm~\ref{alg:concept_fusion} describes our position-aware cross-lingual concept fusion mechanism in detail. Given an English concept chain $C_e$ and a local language concept chain $C_l$, we iteratively integrate local concepts into the fused chain $C_f$ (initialized as $C_e$). For each local concept $c_{t}^{j}$, we compute its embedding $e_t$ and identify the position $k^{*}$ of the most similar concept in the current fused chain using cosine similarity. If the maximum similarity $s_{\max}$ exceeds the threshold $\tau$, we proceed to determine the insertion position. Rather than simply appending the concept, we compare the average similarity of $c_{l}^{j}$ to all concepts positioned before $k^{*}$ ($s_{left}$) versus those after $k^{*}$ ($s_{right}$). This bidirectional context comparison ensures that the inserted concept is positioned where it exhibits the strongest semantic coherence with surrounding concepts, thereby preserving the logical structure and clinical reasoning flow of the chain.
\begin{algorithm}
\caption{Position-Aware Concept Fusion}
\label{alg:concept_fusion}
\SetAlgoLined
\KwIn{English Concept Chain $C_{e} = \{c_{e}^{1}, \ldots, c_{e}^{n}\}$, Local Concept Chain $C_{l} = \{c_{l}^{1}, \ldots, c_{l}^{m}\}$, Embedding function $f$, Threshold $\tau$}
\KwOut{Fused concept chain $C_{f}$}

$C_{f} \leftarrow C_{e}$\;

$E_{f} \leftarrow \{f(c_{e}^{1}), \ldots, f(c_{e}^{n})\}$\;

\For{$c_{l}^{j} \in C_{l}$}{
    $e_{l} \leftarrow f(c_{l}^{j})$\;
    
    $k^{*} \leftarrow \text{argmax}_{k \in [1,|C_{f}|]} \cos(e_{l}, E_{f}^{k})$\;
    
    $s_{\max} \leftarrow \cos(e_{l}, E_{f}^{k^{*}})$\;
    
    \If{$s_{\max} \geq \tau$}{
        $s_{left} \leftarrow \frac{1}{k^{*}-1} \sum_{i=1}^{k^{*}-1} \cos(e_{l}, E_{f}^{i})$\;
        $s_{right} \leftarrow \frac{1}{|C_{f}|-k^{*}} \sum_{i=k^{*}+1}^{|C_{f}|} \cos(e_{l}, E_{f}^{i})$\;
        \eIf{$s_{left} > s_{right}$}{
            $p \leftarrow k^{*}$\;
        }{
            $p \leftarrow k^{*}+1$\;
        }
        
        Insert $c_{l}^{j}$ into $C_{f}$ at position $p$\;
        Insert $e_{l}$ into $E_{f}$ at position $p$\;
    }
}

\textbf{return} $C_{f}$\;
\end{algorithm}

\section{Overall Results on \evalName}
\label{sec:full-results}

\paragraph{Overall Performance.} Table~\ref{tab:medx-results} summarizes results on \evalName across languages and tasks. \sysName achieves the best or near-best performance across all languages, outperforming strong baselines on long-form QA metrics, including Overall, Correctness, Completeness, and Pass Rate, while maintaining high Safety and low Hallucination. Consistent gains in NLI accuracy further demonstrate the effectiveness of cross-lingual co-reasoning and knowledge grounding for reliable multilingual medical decision-making.

\paragraph{Cross-lingual and Low-resource Analysis.}
A closer look reveals that the advantages of \sysName are particularly pronounced in low-resource languages such as Swahili, Yoruba, and Zulu. Compared to direct prompting, \sysName yields larger improvements in completeness, hallucination control, and pass rate, effectively narrowing the performance gap between English and underrepresented languages. This trend highlights the effectiveness of pivot-anchored co-reasoning in preserving logical structure while incorporating localized clinical knowledge, leading to more robust and equitable multilingual medical reasoning.

\begin{table*}
\resizebox{\textwidth}{!}{
\centering
\begin{tabular}{lcccccccc}
\toprule
\multirow{2}{*}{\textbf{Language}} & \multirow{2}{*}{\textbf{Model}} & \multicolumn{6}{c}{\textbf{Long-form QA}} & \textbf{NLI} \\
\cmidrule(lr){3-8} \cmidrule(lr){9-9}
& & \textbf{Overall} & \textbf{Correctness} & \textbf{Completeness} & \textbf{Safety} & \textbf{Hallucination} & \textbf{Pass Rate} & \textbf{Accuracy (\%)} \\
\midrule
\multirow{5}{*}{EN} 
& GPT-5.2 & 4.415 & 4.455 & 4.615 & 4.815 & 4.410 & 0.900 & 76.67 \\
& GPT-5.1 & 4.425 & 4.485 & 4.590 & \textbf{4.880} & 4.385 & 0.915 & 76.67 \\
& GPT-4o & 4.210 & 4.270 & 4.245 & 4.845 & 4.380 & 0.855 & \textbf{78.67} \\
& Claude-3.5-haiku & 4.200 & 4.240 & 4.265 & 4.830 & 4.305 & 0.870 & 69.33 \\
& Ours (GPT-5.1) & \textbf{4.600} & \textbf{4.640} & \textbf{4.850} & 4.860 & \textbf{4.460} & \textbf{0.939} & 77.33 \\
\midrule
\multirow{5}{*}{ZH} 
& GPT-5.2 & 4.420 & 4.460 & 4.605 & 4.800 & 4.360 & 0.915 & 74.00 \\
& GPT-5.1 & 4.435 & 4.490 & 4.660 & 4.845 & \textbf{4.395} & 0.905 & 74.67 \\
& GPT-4o & 4.205 & 4.270 & 4.270 & \textbf{4.855} & 4.250 & 0.860 & 74.67 \\
& Claude-3.5-haiku & 4.150 & 4.205 & 4.295 & 4.735 & 4.140 & 0.845 & 69.33 \\
& Ours (GPT-5.1) & \textbf{4.530} & \textbf{4.550} & \textbf{4.890} & 4.780 & 4.390 & \textbf{0.935} & \textbf{76.67} \\
\midrule
\multirow{5}{*}{JP} 
& GPT-5.2 & 4.335 & 4.405 & 4.520 & 4.855 & 4.330 & 0.885 & 76.67 \\
& GPT-5.1 & 4.330 & 4.435 & 4.525 & \textbf{4.865} & 4.330 & 0.895 & 77.33 \\
& GPT-4o & 4.080 & 4.135 & 4.135 & 4.830 & 4.145 & 0.820 & 76.00 \\
& Claude-3.5-haiku & 4.100 & 4.205 & 4.170 & 4.860 & 4.095 & 0.845 & 68.67 \\
& Ours (GPT-5.1) & \textbf{4.430} & \textbf{4.500} & \textbf{4.850} & 4.800 & \textbf{4.330} & \textbf{0.930} & \textbf{78.00} \\
\midrule
\multirow{5}{*}{KO} 
& GPT-5.2 & 4.410 & 4.460 & 4.655 & \textbf{4.815} & 4.325 & 0.915 & 64.67 \\
& GPT-5.1 & 4.390 & 4.475 & 4.610 & 4.795 & 4.330 & 0.890 & 62.67 \\
& GPT-4o & 4.055 & 4.130 & 4.105 & 4.805 & 4.225 & 0.840 & 61.33 \\
& Claude-3.5-haiku & 4.115 & 4.170 & 4.240 & 4.795 & 4.110 & 0.860 & 60.00 \\
& Ours (GPT-5.1) & \textbf{4.540} & \textbf{4.570} & \textbf{4.840} & 4.790 & \textbf{4.450} & \textbf{0.920} & \textbf{67.33} \\
\midrule
\multirow{5}{*}{SW} 
& GPT-5.2 & 4.340 & 4.415 & 4.590 & 4.795 & 4.345 & 0.890 & 70.00 \\
& GPT-5.1 & 4.330 & 4.385 & 4.545 & 4.805 & 4.255 & 0.885 & 72.67 \\
& GPT-4o & 4.040 & 4.070 & 4.115 & 4.755 & 4.155 & 0.815 & 69.33 \\
& Claude-3.5-haiku & 3.825 & 3.855 & 3.950 & 4.565 & 3.780 & 0.730 & 56.67 \\
& Ours (GPT-5.1) & \textbf{4.550} & \textbf{4.570} & \textbf{4.800} & \textbf{4.820} & \textbf{4.470} & \textbf{0.945} & \textbf{73.33} \\
\midrule
\multirow{5}{*}{TH} 
& GPT-5.2 & 4.440 & 4.490 & 4.605 & 4.835 & 4.430 & 0.900 & 68.00 \\
& GPT-5.1 & 4.490 & 4.535 & 4.645 & 4.830 & 4.545 & 0.910 & 70.00 \\
& GPT-4o & 4.160 & 4.215 & 4.180 & 4.890 & 4.280 & 0.855 & 70.00 \\
& Claude-3.5-haiku & 4.135 & 4.205 & 4.200 & 4.790 & 4.330 & 0.855 & 63.33 \\
& Ours (GPT-5.1) & \textbf{4.660} & \textbf{4.670} & \textbf{4.880} & \textbf{4.910} & \textbf{4.620} & \textbf{0.960} & 71.33 \\
\midrule
\multirow{5}{*}{YO} 
& GPT-5.2 & 4.065 & 4.135 & 4.225 & 4.550 & 4.060 & 0.795 & 62.00 \\
& GPT-5.1 & 4.090 & 4.160 & 4.310 & 4.595 & 4.080 & 0.810 & 63.33 \\
& GPT-4o & 3.290 & 3.325 & 3.400 & 4.095 & 3.205 & 0.455 & 62.67 \\
& Claude-3.5-haiku & 3.545 & 3.555 & 3.650 & 4.435 & 3.585 & 0.600 & 54.67 \\
& Ours (GPT-5.1) & \textbf{4.450} & \textbf{4.460} & \textbf{4.790} & \textbf{4.700} & \textbf{4.230} & \textbf{0.910} & \textbf{68.67} \\
\midrule

\multirow{5}{*}{ZU} 
& GPT-5.2 & 4.210 & 4.280 & 4.440 & 4.715 & 4.195 & 0.860 & 66.00 \\
& GPT-5.1 & 4.160 & 4.230 & 4.445 & 4.695 & 4.145 & 0.850 & 64.00 \\
& GPT-4o & 3.780 & 3.805 & 3.875 & 4.605 & 3.885 & 0.725 & 66.00 \\
& Claude-3.5-haiku & 3.670 & 3.705 & 3.865 & 4.505 & 3.685 & 0.645 & 62.00 \\
& Ours (GPT-5.1) & \textbf{4.420} & \textbf{4.710} & \textbf{4.770} & \textbf{4.730} & \textbf{4.310} & \textbf{0.915} & \textbf{68.00} \\
\bottomrule
\end{tabular}
}
\caption{Complete evaluation results across different languages on \evalName.}
\label{tab:medx-results}
\end{table*}

\section{Reasoning Training Data}
The comparative statistics of the MMedBench training set and our dataset are shown in Table~\ref{tab:training_data_statistics}. Using \sysName, we generate forward reasoning by inputting training questions and obtaining corresponding reasoning chains and answers. We only use those instances with correct final answers for training.

\label{sec:training_data}

\begin{table*}
    \centering
    \small
    \begin{tabular}{ccccccc}
    \toprule
         Train Set & Chinese & English & French & Japanese & Russian & Spanish  \\
        \midrule
         MMedBench & 27,400 & 10,178 & 2,171 & 1,590 & 1,052 & 2,656 \\
         \dataName & 8,627 & 9,513 & 1,603 &  1,392 & 846 & 2,487\\
    \bottomrule
    \end{tabular}
    \caption{Training data statistics of MMedBench and \dataName}
    \label{tab:training_data_statistics}
\end{table*}

\section{Full Results on MMedBench}
We include multiple large language models pre-trained specifically for the medical domain on MMedBench for comparison, including BioMistral-7B~\citep{labrak2024biomistral}, MMedLM2-7B~\citep{chen-etal-2025-towards-medical}, and MedGemma-27B~\citep{sellergren2025medgemma}. Full results are reported in Table~\ref{tab:full-mmedbench-result}.
\paragraph{Overall Comparison.} Table~\ref{tab:full-mmedbench-result} compares multilingual performance on MMedBench across medical-domain and general-purpose LLMs. Among domain-specific models, MedGemma-27B achieves the strongest average performance (65.88), outperforming BioMistral-7B and MMedLM2-7B, but still exhibits notable variance across languages, particularly weaker results in French and Japanese. This suggests that medical pre-training alone does not guarantee robust multilingual generalization.

\paragraph{Effect of Training Data and Model Scale.} For general-purpose models fine-tuned on different datasets, training on \dataName consistently improves multilingual performance compared to MMedBench across model scales. In particular, Qwen2.5-14B trained on \dataName achieves the best overall average score (75.97), with clear gains across all non-English languages and especially large improvements in French and Japanese. Similar trends are observed for Qwen2.5-7B and Gemma-7B-it, indicating that \dataName provides more effective cross-lingual medical supervision and that performance gains scale with model capacity.

\begin{table*}
    \centering
    \small
    \setlength{\tabcolsep}{5pt}
    \begin{tabular}{lccccccccc}
        \toprule
        Model & Train Set& \textbf{Chinese} & \textbf{English} & \textbf{French} &
         \textbf{Japanese} & \textbf{Russian} & \textbf{Spanish} & \textbf{Avg.} \\
         \midrule
         BioMistral-7B & - & 25.89 & 19.17 & 10.13 & 8.54 & 54.3 & 25.67 & 24.66 \\
         MMedLM2-7B & - & 70.43 & 58.13 & 54.27 & 38.26 & 71.88 & 64.95 & 59.32 \\
         MedGemma-27B & - & 73.50 & 71.09 & 41.16 & 60.08 & 72.27 & 79.72 & 65.88 \\
        \midrule
        \multirow{2}{*}{Gemma-7B-it}& MMedBench & 56.07 & 52.16 & 34.89 & 34.67 & 63.28 & 57.51 & 49.45 \\
        & \dataName & 52.38 & 52.08 & 40.03 & 41.71 & 64.45 & 58.06 & 51.39 \\
        \midrule
        \multirow{2}{*}{Qwen2.5-7B}& MMedBench & 81.47 &61.67 & 47.72 & 48.74 & 69.53 & 67.18 & 62.16 \\
        & \dataName & 78.78 &62.37 & 54.66 & 56.78 & 70.31 & 69.22 & 64.98 \\
        \midrule
        \multirow{2}{*}{Qwen2.5-14B} &MMedBench & \textbf{84.47} &71.48 & 64.15 & 66.33 & 75.39 & 78.77 & 73.11 \\
        & \dataName & 82.89 & \textbf{75.73} & \textbf{72.03} & \textbf{68.34} & \textbf{75.78} & \textbf{82.39} & \textbf{75.97} \\
        \bottomrule
    \end{tabular}
    \caption{Performance comparison across languages on MMedBench.}
    \label{tab:full-mmedbench-result}
\end{table*}

\section{Expert Evaluation.}
\label{sec:expert_evaluation}

We recruit the expert physicians through social media. For the reasoning quality assessment experiment, we randomly sample questions in Japanese, Spanish, Chinese, and Korean from the MMLU-ProX benchmark, and generate reasoning and answers using GPT-5.1 and \sysName. In the first task, we retain only the cases where both models produce correct answers, resulting in 40 question–answer pairs. For the labeling task, we random one question for evaluation. The guidelines provided to physician experts are shown in Figure~\ref{box:expert_guidelines} and the example pairwise evaluation for first task is shown in Table~\ref{tab:pairwise_example}.

\section{Implementation Details}
\label{sec:detailed_implementation}
We provide all hyperparameters and experimental settings in this section.

\paragraph{Prompts.} For parallel reasoning and concept extraction, we use the prompts shown in Figures~\ref{box:reasoning_prompt} and~\ref{box:extraction_prompt}. Final answer generation is performed using the prompt in Figure~\ref{box:answer_generation_prompt}. For LLM-as-a-judge evaluation in long-form QA, we adopt the system prompt in Figure~\ref{box:judge_system_prompt} together with the evaluation prompt in Figure~\ref{box:judge_eval_prompt}.

\paragraph{Knowledge Retrieval.} We construct language-specific medical knowledge bases from MSD Manuals and AFRIDOC-MT. Detailed statistics of the documents for each language are reported in Table~\ref{tab:knowledge_base}. Given a query in a particular language, we retrieve relevant documents from the corresponding language-specific knowledge base. We use BGE-M3 as the retriever and reranker, retrieving the top 10 documents in the initial retrieval stage and reranking the top 3 documents for final use.

\begin{table*}
    \centering
    \small
    \begin{tabular}{ccccccccccc}
    \toprule
         EN & ZH & JA & KO & DE & FR & ES & IT & SW & YO & ZU  \\
    \midrule
         2,441 & 2,857 & 2,502 & 3,428 & 2,855 & 3,044 & 2,943 & 2,960 & 3,491 & 1,148 & 1,148 \\
    \bottomrule
    \end{tabular}
    \caption{Document statistics of multilingual knowledge base.}
    \label{tab:knowledge_base}
\end{table*}

\begin{figure*}
\begin{tcolorbox}[title=\textbf{Pairwise Comparison Guidelines}, colback=theme_purple!20,
colframe=theme_purple,
coltitle=black,
boxrule=0.5pt,
left=3mm, right=3mm, top=2mm, bottom=2mm]
\textbf{Introduction}: We want to verify which “reasoning” path is more reasonable. Here, reasoning refers to the structured sequence of diagnostic or decision-making steps that link clinical evidence to a conclusion, analogous to clinical reasoning in medical practice.

\textbf{Task 1: Pairwise Comparison}

\textbf{Instruction:}

You will be shown a clinical question and two reasoning explanations (A and B) for the same case. Do not judge only based on the final answer, but focus on the reasoning quality.

\textbf{Please evaluate which is better (or tie) considering the following dimensions:}
\begin{itemize}
    \item \textbf{Reasoning Clarity}: Which reasoning or explanation is more logically organized?
    \item \textbf{Clinical Soundness}: Which reasoning is more medically reasonable?
    \item \textbf{Hallucination/Safety}: Which reasoning is less likely to mislead clinical judgement?
    \item \textbf{Local Clinical Naturalness}: Which reasoning sounds more natural in the local clinical context?
\end{itemize}

Please select A, B, or Tie in each evaluation dimension.

\textbf{Task 2: Label correctness and incorrectness}

For each text, please highlight:
\begin{itemize}
    \item \textcolor{green}{Green}: sentences or phrases that are clinically correct/appropriate (valid medical statements, reasonable clinical inferences).
    \item \textcolor{red}{Red}: sentences or phrases that are clinically incorrect, unsupported, or potentially unsafe/misleading.
\end{itemize}
\textbf{Example}:\\
Los azúcares libres, según la OMS, \textcolor{red}{incluyen los monosacáridos} y disacáridos añadidos a los alimentos y bebidas, \textcolor{green}{además de los azúcares presentes} de forma natural en miel, 

\end{tcolorbox}
\caption{Physician expert pairwise comparison guidelines.}
\label{box:expert_guidelines}
\end{figure*}

\begin{table*}
\renewcommand{\arraystretch}{1.2}
\small
\centering
\begin{tabular}{>{\columncolor{theme_blue!50}\raggedright\arraybackslash}m{3cm} | 
>{\columncolor{theme_purple!50}\raggedright\arraybackslash}m{6cm} | 
>{\columncolor{theme_purple!50}\raggedright\arraybackslash}m{6cm}}
\toprule
\textbf{Question} & \textbf{A\_Reasoning} & \textbf{B\_Reasoning} \\
\midrule
Observa una lesión en la mucosa alveolar labial del incisivo lateral inferior derecho de uno de sus pacientes y decide tomar una biopsia para enviar al informe de patología oral. ¿Qué nervio requeriría anestesia local para realizar una biopsia sin dolor?

\textit{Options:} A: El nervio mentoniano; B: El nervio hipogloso; C: El nervio incisivo; D: El nervio bucal; E: El nervio alveolar inferior; F: El nervio infraorbitario; G: El nervio facial; H: El nervio maxilar; I: El nervio lingual; J: El nervio trigémino
& 
La lesión se localiza en la mucosa alveolar labial del incisivo lateral inferior derecho. La sensibilidad de la mucosa y piel del labio inferior y región del mentón está dada por el nervio mentoniano, que es la rama terminal del nervio alveolar inferior tras salir por el foramen mentoniano. El nervio incisivo inerva principalmente los dientes anteriores (pulpa y tejidos periapicales) pero no la mucosa labial. Por tanto, para una biopsia de la mucosa alveolar labial en esa zona, el nervio que debe anestesiarse específicamente es el nervio mentoniano.
& 
La lesión se localiza en la mucosa alveolar labial a nivel del incisivo lateral inferior derecho. La sensibilidad de la mucosa/encía vestibular (labial) de los incisivos, canino y primer premolar mandibulares está dada por el nervio mentoniano, que es la rama terminal del nervio alveolar inferior. 

En el contexto aportado se describe que el nervio alveolar inferior, rama del nervio mandibular (V3), discurre dentro de la mandíbula y da una rama terminal, el nervio mentoniano, que emerge por el foramen mentoniano en la región de los premolares e inerva el labio inferior, el mentón y la mucosa/gingiva lateral que recubre los dientes incisivos inferiores, canino y primer premolar. 

Tras emitir el nervio mentoniano, el nervio alveolar inferior continúa hacia anterior como nervio incisivo, que inerva los dientes y el hueso anteriores al foramen mentoniano, pero no la mucosa labial externa. 

Por lo tanto, para realizar una biopsia indolora de la mucosa alveolar labial del incisivo lateral inferior, debe anestesiarse el nervio mentoniano, no el nervio incisivo ni el alveolar inferior de forma proximal. 

Las demás opciones corresponden a nervios con territorios de inervación diferentes (lingual, bucal, infraorbitario, maxilar, facial, hipogloso, trigémino completo) y no son los responsables de la sensibilidad en esa zona específica.
\\
\midrule
La mayoría de los ácidos grasos libres se transportan en la sangre:
Options: A: dentro de los glóbulos rojos.; B: combinados con oxígeno.; C: como triglicéridos.; D: como moléculas libres.; E: unidos a los glóbulos blancos.; F: como lipoproteínas.; G: unidos a la hemoglobina.; H: combinados con glucosa.; I: en forma de colesterol.; J: unidos a la albúmina.

&
Los ácidos grasos libres, una vez liberados del tejido adiposo al torrente sanguíneo, son moléculas hidrofóbicas que no se disuelven bien en el plasma acuoso. Para poder transportarse de forma eficiente y segura en la sangre, se unen de manera no covalente principalmente a la albúmina sérica. No viajan como triglicéridos, colesterol ni como parte de lipoproteínas; estas formas se utilizan sobre todo para el transporte de lípidos más complejos (triglicéridos y colesterol esterificado), no de ácidos grasos libres. Tampoco se unen a eritrocitos, leucocitos, hemoglobina, glucosa ni oxígeno. Por lo tanto, la forma principal de transporte de los ácidos grasos libres en la sangre es unidos a la albúmina.
& 
Los ácidos grasos libres (no esterificados) son moléculas hidrofóbicas, por lo que no pueden circular libremente disueltas en el plasma, que es un medio acuoso. Para poder ser transportados en sangre, deben unirse a proteínas transportadoras. En el caso específico de los ácidos grasos libres, el principal transportador en el plasma es la albúmina sérica, que posee sitios de unión para estos ácidos grasos y los lleva desde el tejido adiposo hacia otros tejidos (como músculo e hígado) donde se utilizan como fuente de energía. No se transportan mayoritariamente dentro de eritrocitos, ni unidos a hemoglobina, leucocitos, oxígeno, glucosa, ni formando triglicéridos, colesterol o lipoproteínas (estas estructuras transportan principalmente triglicéridos y colesterol, no los ácidos grasos libres como tales). Por tanto, la forma principal de transporte de los ácidos grasos libres en sangre es unidos a la albúmina.

\\
\bottomrule
\end{tabular}
\caption{Example of pairwise reasoning comparison in Spanish. In both cases, A\_reasoning corresponds to the GPT-5.1 baseline, while B\_reasoning represents reasoning generated by \sysName with GPT-5.1 as the backbone.}
\label{tab:pairwise_example}
\end{table*}

\begin{figure*}
\begin{tcolorbox}[
title=\textbf{Parallel Reasoning Prompt},
colback=theme_purple!20,
colframe=theme_purple,
coltitle=black
]
Your task is to assist healthcare professionals in clinical reasoning by providing well-thought-out answers to medical questions.
Please first think step by step using the {language} language and then provide your final answer. Your response will be used for research purpose only, so please provide a definite answer (e.g., A, B, C, or D).

**Question**:\\
\{question\}

**Options**:\\
\{options\}

**Output Format: **

Please provide your reasoning process in a step-by-step manner using \{language\} language, followed by your final answer. Use the following format:

\{\{
    "reasoning": "Your detailed reasoning process here",
    "answer": "Your definite answer here, e.g., A, B, C, or D"
\}\}

\end{tcolorbox}
\caption{Reasoning Prompt}
\label{box:reasoning_prompt}
\end{figure*}

\begin{figure*}
    
\begin{tcolorbox}[
title=\textbf{Concept Extraction Prompt},
colback=theme_purple!20,
colframe=theme_purple,
coltitle=black
]
Your task is to transform a reasoning trace into a concise, ordered **concept chain**.

**Definitions**

- "A concept" is an atomic, reusable medical idea or clinical finding that contributes to clinical reasoning. (e.g., "chronic cough", "chest X-ray", "smoker's history")

- "Concept chain" is an ordered list of these concepts that follows the original reasoning order.

**Instructions**

- Read the reasoning trace carefully.

- Extract key concepts in the same order as they appear

- Each concept must be concise and represent only one idea

- Prefer clinical or scientific terms over long sentences

- Do not invent new concepts that are not implied by the reasoning

- Merge duplicates or near-duplicates into one concept, but keep the order consistent

- The concept chain should be as short as possible while capturing all essential reasoning 
steps, and must not include or expose any answer options.

- Keep the output in the same language as the reasoning.

**Output Format (plain text, no explanation):**

Provide the concept chain as a list in the following format:

$["Concept 1", "Concept 2", "Concept 3", ...]$

Now process the following reasoning trace and output only the concept chain in \{language\} language.

\{reasoning\_trace\}

\end{tcolorbox}
\caption{Concept Extraction Prompt}
\label{box:extraction_prompt}
\end{figure*}

\begin{figure*}
    
\begin{tcolorbox}[
title=\textbf{Final Answer Generation Prompt},
colback=theme_purple!20,
colframe=theme_purple,
coltitle=black
]
Your task is to generate a final clinical answer for a multi-option question by integrating a **concept reasoning chain** with **the retrieved medical documents** from the concept chain and your **prior medical knowledge**.

**Inputs**

- Question:

\{question\}

- Options:

\{options\}

- Concept Reasoning Chain (in order):

\{concept\_chain\}

- Referenced Context (mainly contains multiple documents, guidelines, or passages):

\{context\}

**Instructions**

1. First, carefully read the concept reasoning chain. Treat it as a DRAFT reasoning path, not as guaranteed truth.

2. Then, carefully read the referenced context. Use it to VERIFY, CORRECT, or REFINE the reasoning chain.

3. Use ONLY information that is supported by the referenced context and widely accepted medical knowledge. DO NOT directly mention the concept chain. ORGANIZE your reasoning in a clear, logical manner.

4. Finally, select the MOST APPROPRIATE option as your final answer based on the verified and refined reasoning.

5. Output the reasoning in {language}, regardless of the input language.

**Output Format**:

Return VALID JSON ONLY, following this format:
\{\{
    "reasoning": "Your verified and refined reasoning process here",
    "answer": "Your final answer here, e.g., A, B, C, or D"
\}\}

\end{tcolorbox}
\caption{Final Answer Generation Prompt}
\label{box:answer_generation_prompt}
\end{figure*}

\begin{figure*}
    
\begin{tcolorbox}[
title=\textbf{Judge System Prompt},
colback=theme_purple!20,
colframe=theme_purple,
coltitle=black
]
You are an objective and rigorous evaluator for medical question answering. \\
You will be given: \\
- a Question\\
- a Ground-Truth Answer (reference)\\
- a Model Answer (candidate) \\
Your task is to evaluate the Model Answer relative to the Ground-Truth Answer. \\
Evaluation principles: \\
Prioritize factual correctness, clinical safety, and alignment with the reference.\\
Do NOT penalize harmless extra details if they are correct and do not contradict the reference. \\
Penalize contradictions, fabricated facts, or unsafe medical advice. \\
If the reference is brief but the model answer is longer, judge consistency and medical plausibility. \\
If a detail cannot be verified from the reference, treat it as uncertain rather than incorrect.\\
Output MUST be valid JSON only
\end{tcolorbox}
\caption{The system prompt of LLM-as-a-judge in the evaluation of long-form QA task.}
\label{box:judge_system_prompt}
\end{figure*}

\begin{figure*}
\begin{tcolorbox}[
title=\textbf{Judge Evaluation Prompt},
colback=theme_purple!20,
colframe=theme_purple,
coltitle=black
]
Question:\\
\{question\}\\

Ground-Truth Answer:\\
\{gold\}\\

Model Answer:\\
\{pred\}\\

Return JSON with EXACTLY the following fields and no others: \\
\{\{ \\
"overall\_score": 1-5,\\
"correctness": 1-5,\\
"completeness": 1-5,\\
"safety": 1-5,\\
"hallucination": 1-5\\
\}\}

Scoring rules:\\
- 5 = excellent\\
- 4 = good with minor issues\\
- 3 = partially correct or incomplete\\
- 2 = major issues\\
- 1 = mostly incorrect or unsafe\\

For hallucination:\\
- 5 = no hallucination\\
- 3 = some uncertain additions\\
- 1 = clear hallucinations or fabricated facts\\
\end{tcolorbox}
\caption{The evaluation prompt of LLM-as-a-judge in the evaluation of long-form QA task.}
\label{box:judge_eval_prompt}
\end{figure*}


\end{document}